# Salient Local 3D Features for 3D Shape Retrieval

Afzal Godil[a], Asim Imdad Wagan[b]
[a]National Institute of Standards and Technology, Gaithersburg, MD, USA
[b]Dept of CSE, QUEST University, Nawabshah, Pakistan

## ABSTRACT

In this paper we describe a new formulation for the 3D salient local features based on the voxel grid inspired by the Scale Invariant Feature Transform (SIFT). We use it to identify the salient keypoints (invariant points) on a 3D voxelized model and calculate invariant 3D local feature descriptors at these keypoints. We then use the bag of words approach on the 3D local features to represent the 3D models for shape retrieval. The advantages of the method are that it can be applied to rigid as well as to articulated and deformable 3D models. Finally, this approach is applied for 3D Shape Retrieval on the McGill articulated shape benchmark and then the retrieval results are presented and compared to other methods.

## 1. INTRODUCTION

With recent advances in 3D modeling programs and 3D scanning technologies, large numbers of 3D models are created and are widely used in many diverse fields such as, computer graphics, computer aided design and manufacturing, computer vision, entertainment and games, cultural heritage, medical imaging, structural biology, and other fields. This has created an impetus to develop effective 3D shape retrieval algorithms for these domains and has made the field of 3D shape retrieval become an active area of research. One of the main areas of research in 3D Shape retrieval is in the field of shape descriptors (feature descriptors). Several methods have been proposed for different types of shape descriptors, such as view-based [1, 14, 22], statistics-based [2], transform-based [3]. Shape descriptors can also be classified as local and global shape descriptors. More details about the different methods can be found in the following review papers [6, 23]. Most of the retrieval methods described above, except the local feature based methods, perform poorly for non-rigid models. Lots of natural and manmade objects have articulation and deformation, such as: humans and animals in different poses; and deformation in protein molecules, etc. Hence comparing non-rigid 3D models with articulation and deformation is still a very difficult and very challenging problem.

One of the main benchmark for non-rigid models is the McGill 3D shape benchmark [9, 20], which includes models with both deformation and articulated parts. For the non-rigid 3D shape retrieval, Ohbuchi et. al. [5, 7] has applied salient local features to render depth images from a number of views to 3D models and then by applying the "bag-of-words" approach. Jain et al. [17] has applied the spectral approach to the retrieval of articulation 3D models. Mahmoudi et al. [11] has discussed number of different signatures for 3D models with articulation. Elad et al. [12] has proposed extracting bending-invariant signatures by applying multidimensional scaling method. Li et al. [10] applied spin images along with bag-of-words approach to the retrieval of 3D articulation models. Lian et al. [13] applied multidimensional scaling method to 3D models and then applied bag-of-words approach to salient local features of the rendered depth images. Finally, a track on non-rigid 3D shape retrieval was organized under 3D shape retrieval contest (SHREC 2010) which was held at the EuroGraphics Workshop on 3D Object Retrieval [21].

We propose a new formulation for the 3D salient local features based on the voxel grid inspired by the Scale Invariant Feature Transform (SIFT) [4] and apply it to the field of non-rigid 3D Model retrieval. The advantage of local features is that the method can be applied to both rigid models as well as to articulated and deformable 3D models. The local features also have been effectively used for partial shape matching [10].

There are a number of other independent implementation of the 3D SIFT method that we are aware off. The first one is by Scovanner et al. [15], for the application to activity recognition for video analysis. The second implementation was developed by Cheung et al. [16], for the application of medial volumetric images analysis. The third implementation is by Ohbuchi et al. [8] for 3D shape retrieval. Our 3D salient local feature implementation is specifically developed for 3D models and hence the salient keypoints only exist on the surface and the orientation normalization is completely 3D and is based on the surface normal at the salient keypoints. Also the orientation histogram is calculated on a geodesic sphere so that the triangles are uniformly distributed and are equal in size. Also our implementation is much simpler and faster than other 3D SIFT implementations.

## 1.1 Contributions and Outline

Our main contribution is to develop a 3D salient local feature implementation only for 3D models and hence the salient keypoints only exist on the surface and the orientation normalization is completely 3D and is based on the surface normal at the salient keypoints. Also the orientation histogram is calculated on a geodesic sphere so that the triangles are uniformly distributed and are equal in area. We then use the bag of words approach on the 3D local features to represent the 3D models for shape retrieval. Also our implementation is much simpler and faster than other 3D SIFT implementations is specifically developed for 3D model retrieval.

The paper is organized as follows: In Section 2, we present more details about the algorithm, the voxelization, scale-space for voxels, and identify the extrema points. In Section 3, we describe the feature vectors and orientation normalization. We then present experimental results in Section 5, and conclude in Section 6.

## 2. ALGORITHM

In the following subsections we will present more details about the algorithm.

### 2.1 Voxelization

During the voxelization the mesh is converted into voxels by visualizing the mesh on the plane which passes through the mesh parallel to its main axis. For voxelization, we have used the 3D Voxelizer matlab code [18].

### 2.2 Scale-Space for Voxels

Scale space for 3D space can be constructed by applying 3D Gaussian filters at increasingly large scales by changing the value of k for each scale. It is similar to what is done for images in the SIFT algorithm but for 3 dimensions (Figure 1.).

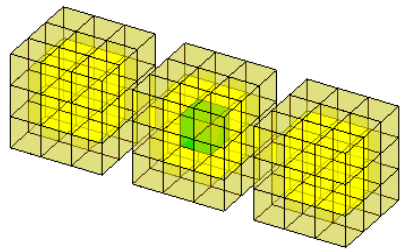

Figure 1. Shows the Scale space for 3D space

If we describe the model by M(x,y,z) then the formula for each layer can be represented by:

$$M_k = M(x,y,z) * G(x,y,z,k\delta)$$

$$G(x,y,z,k\delta) = \frac{1}{(\sqrt{2\pi}k\delta)^3} e^{-(x^2+y^2+z^2)/2(k\delta)^2}$$

By taking the difference of the Gaussian model at each corresponding layer we can find out the points which give consistently better response. These voxels are further processed to find the voxels that give the highest response.

$$DoG_k = M_k(x,y,z) - M(x,y,z)$$

### 2.3 Maxima Detection and Refinement of Features

To identify the extrema points (salient keypoints) we follow the process which was used in the 2D SIFT algorithm. The model is convolved with the 3 Dimensional Gaussian filter and each convolution represents the scale of the model. The Difference of Gaussian is calculated by subtracting the original model from other scaled model. This gives us the voxels in the space that has correspondingly non invariant information related to them. Figure 2, shows salient keypoints detected on four models.

The extrema points are detected by searching the Difference of Gaussian (DoG) space in all directions. A point is an extrema point if and only if it's a maximum or minimum value compared to all its neighbors and at all the Gaussian scaled models. Finally, we only select the extrema points which are located on the surface of 3D object, since our 3D models are solid. Hence our implementation is not suitable for medical imaging and other video applications.

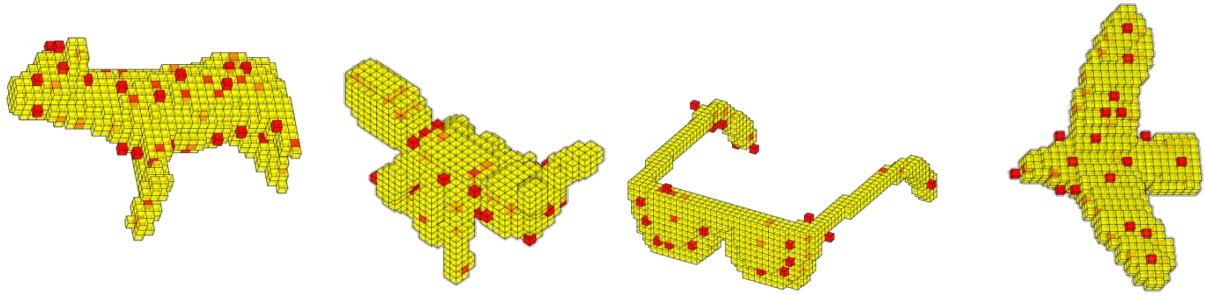

Figure 2. Show four models with detected salient keypoints.

## 3. FEATURE VECTOR

After the detection of the extrema, the models are processed to find the feature vectors for each extrema. This feature vector will represent the model during the detection process. Feature vectors for the extrema points are calculated in a simplified manner compared to the standard SIFT method. An 8x8x8 segment with the keypoint (extrema) at the center of the model is taken as the location of the descriptor. This part is further broken into 8 further subdivisions with each having the size of 4x4x4. Each voxel is then processed to find the orientation and the length of the vector at that position.

### 3.1 Orientation Normalization

When we process each voxel in the subdivision, the orientation is normalized by subtracting the angle of the normal to the surface at that keypoint, this makes the feature vector rotation invariant. Now the angles which are found at the voxel are quantized and a histogram is created based on these angles. The length of the vector is summed at each corresponding quantized value.

$$R = \sqrt{x^2 + y^2 + z^2}$$

$$S = \sqrt{x^2 + y^2}$$

$$\varphi = ar\cos(z/R)$$

$$\theta = ar\sin(y/S) \quad or \quad \pi - \arcsin(y/S)$$

We have created the orientation histogram based on a geodesic sphere [13][14] so that all of the triangles are uniformly distributed and are equal in size. We create a geodesic sphere by subdividing a regular octahedron to obtain finer geodesic spheres with more vertices. In this work, we have only used the third level of geodesic spheres octahedron with 66 vertices. Figure 3, shows some examples of geodesic spheres.

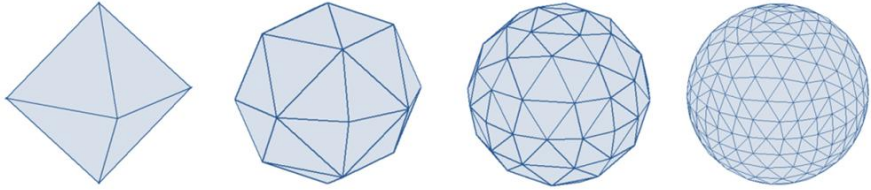

Figure 3. The Geodesic Spheres

## 4. EXPERIMENTAL RESULTS

The experimental results are described in the following subsections.

### 4.1 Benchmark

We conducted our experiments on the database of McGill 3D shape benchmark (Figure 4.). The database consists of a training set with 258 models in 10 classes. We used nearest neighbor (NN), first tier (FT), second tier (ST), and discounted cumulative gain (DCG) as measures of retrieval performance.

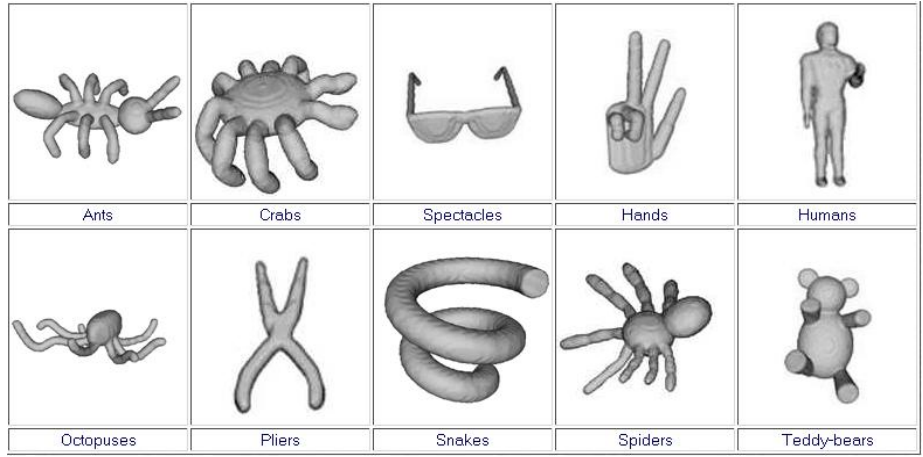

Figure 4. Shows the first object from each class for objects with articulation

### 4.2 Keypoint Matching

The size of the local feature vector at the keypoint depends on the number of vertices selected on the geodesic sphere. After the calculation of the feature vector at the keypoints, all feature vectors for all models are collected. The collection of feature vectors are then clustered using the K-means algorithm. The number of clusters is predefined according to the size of the codebook required.. The number of iterations usually chosen in a way to stop when the difference between K-means cycles the difference of mean error between two cycles is less than a certain constant, but in this study we selected the number of iterations to be 20. The codebook is then used to generate a histogram which is the combination of occurrences of the same type of feature vectors occurring repeatedly.

The generated histogram is a representation of the 3D model. The histogram can be used also to calculate the distance between two 3D models. In our case we calculate the Euclidean distance between two histograms as the final distance between two 3D models. We also used the code based on the SIFT matching, which took the feature descriptors at the keypoints for matching, and Figure 5, shows the matching of key points between pair of four models from the same class.

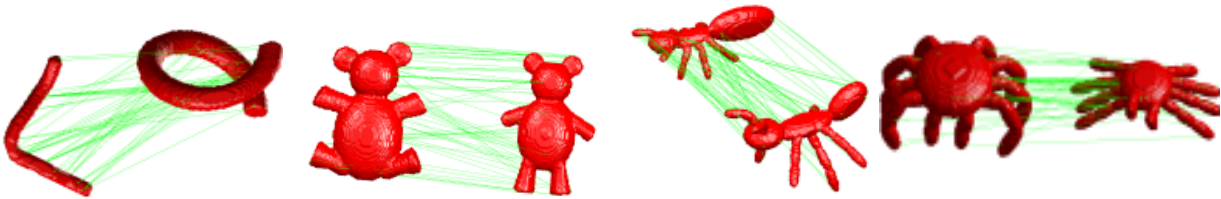

Figure 5. Shows the matching of the key points of four pairs of two different models from the same class.

### 4.3 Evaluation Results

In this section, we present the performance evaluation results of our method applied to the articulated McGill Shape Benchmark. In response to a given set of query objects, an algorithm searches the data-set and returns an ordered list of responses called the ranked list(s). Different evaluation metrics measure different aspects of shape retrieval system. We have employed the following five evaluation measures: Nearest Neighbor (NN), First Tier (FT), Second Tier (ST), Discounted Cumulative Gain (DCG) and Precision Recall Curve. The definitions and implications of these measures can be found in the work of Shilane et al. [19].

Table 1 shows the retrieval statistics (Nearest Neighbor (NN), First Tier (FT), Second Tier (ST), and Discounted Cumulative Gain (DCG) of our 3D local runs with two different parameters. The parameters are: A) feature vector length = 256 and codebook size = 3000; and B) the feature vector size = 1024 and codebook size is 3000. The results from

Obauchi's SIFT-BW method and LFD are also presented. Our results are better than the LFD method but results based on Obauchi's SIFT-BW method are better than our 3D local method. The results are comparable to other widely used methods.

Table 1. The retrieval statistics (Nearest Neighbor (NN), First Tier(FT), Second Tier (ST), and Discounted Cumulative Gain (DCG) are shown.

| Method | NN | FT | ST | DCG |
| --- | --- | --- | --- | --- |
| 3D Local-A | 0.972 | 0.658 | 0.784 | 0.921 |
| 3D Local-B | 0.952 | 0.624 | 0.748 | 0.876 |
| Obauchi (SIFT-BW) | 0.974 | 0.747 | 0.870 | 0.937 |
| LFD | 0.910 | 0.528 | 0.697 | 0.839 |

Figure 6, shows visual results for the spider's class of the articulated shape retrieval database. Only the top 18 results are listed here, in which the green bold framed shape with a question mark is the query shape, and the red bold framed with the X mark shape is the false recall, and only one shape doesn't belong to the spider class. The yellow frame defines the correct class result.

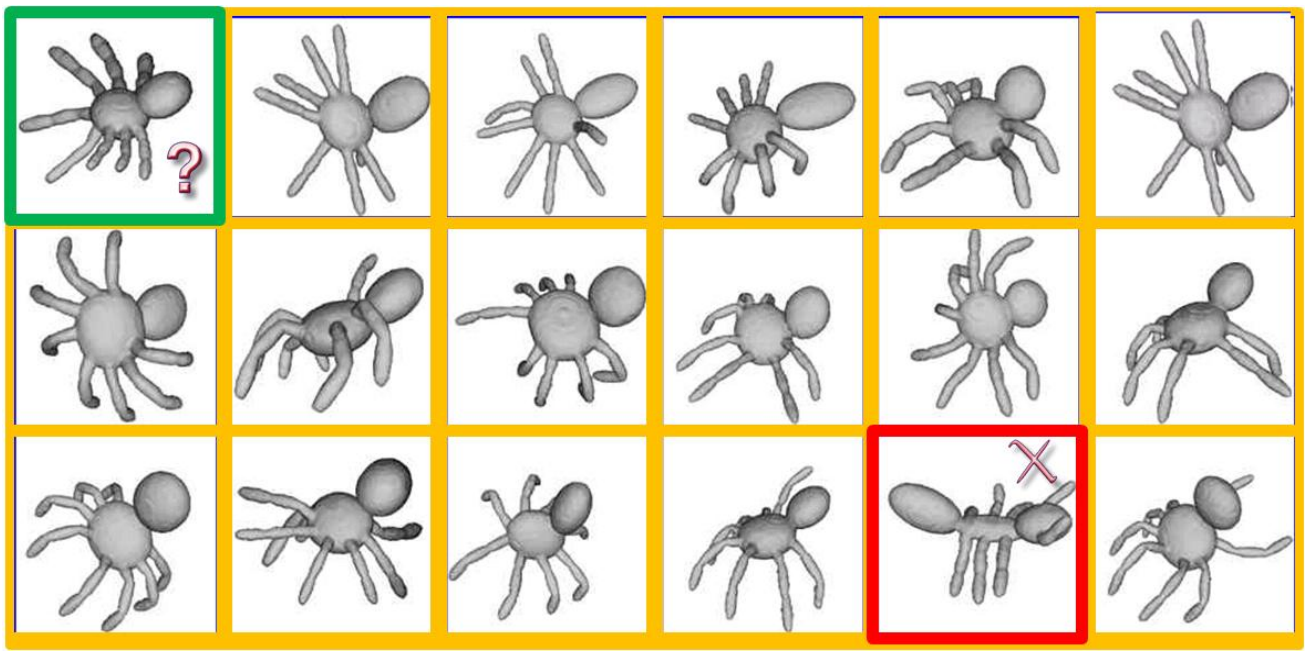

Figure 6. Retrieval results for the spider's class from the McGill database [9, 20]: green bold frame with a question mark defines the query shape, the yellow frame defines the correct class result and the red bold frame with a X mark defines the false pick up.

## 5. CONCLUSION

We have presented a new formulation for the 3D salient local features based on the voxel grid inspired by the SIFT method. We used it to identify the invariant keypoints on a 3D voxelized model and calculate invariant 3D feature descriptors and apply it to the field of non-rigid 3D Model retrieval. The reason for proposing another variation was to create a 3D local feature which is specifically tailored for 3D shape retrieval. Finally, experiments on the McGill articulated shape benchmark are presented and the results are comparable to other widely used methods.

## 6. ACKNOWLEDGMENTS

This work has been supported by the SIMA program. The code for generating the geodesic spheres octahedron was provided by Helin Dutagaci.

## REFERENCES

1. D.Y. Chen, M. Ouhyoung, X.P. Tian, Y.T. Shen. On visual similarity based 3D model retrieval. Eurographics '03, pp. 223-232.
2. R. Osada, T. Funkhouser, B. Chazelle, and D. Dobkin, "Shape distributions," ACM TOG, vol. 21, no. 4, pp. 807–832, 2002
3. M. Kazhdan, T. Funkhouser, and S. Rusinkiewicz, "Rotation invariant spherical harmonic representation of 3D shape descriptors," in Proc. SGP, 2003, pp. 156–164
4. David G. Lowe, Distinctive Image Features from Scale-Invariant Keypoints, Int'l Journal of Computer Vision, 60(2), (2004).
5. R. Ohbuchi, K. Osada, T. Furuya, T. Banno, Salient local visual features for shape-based 3D model retrieval, Proc. SMI '08, (2008).
6. J.W. Tangelder and R. C. Veltkamp, "A survey of content based 3D shape retrieval methods," Multimedia Tools and Applications, 441–471, 2008
7. T. Furuya, R. Ohbuchi, Dense sampling and fast encoding for 3D model retrieval using bag-of-visual features, CIVR '09 Proceeding of the ACM International Conference on Image and Video Retrieval, 2009
8. K Osada, T Furuya, R. Ohbuchi, SHREC'08 Entry: Local Volumetric Features for 3D Model Retrieval, SMI '08, (2008).
9. J. Zhang, K. Siddiqi, D. Macrini, A. Shokoufandeh & S. Dickinson, Retrieving Articulated 3-D Models Using Medial Surfaces and their Graph Spectra. International Workshop On Energy Minimization, 2005.
10. X. Li, A. Godil, EXPLORING THE BAG-OF-WORDS METHOD FOR 3D SHAPE RETRIEVAL, IEEE ICIP 2009
11. M. Mahmoudi and G. Sapiro, "Three-dimensional point cloud recognition via distributions of geometric distances,"Graphical Models, vol. 71, 2009.
12. A. Elad and R. Kimmel, "On bending invariant signatures for surface," PAMI, 2003.
13. Z. Lian, A. Godil, X. Sun, H. Zhang, Non-rigid 3D Shape Retrieval Using Multidimensional Scaling and Bag-of-Features, International Conference on Image Processing (ICIP 2010).
14. H. Dutagaci, A. Godil, B. Sankur, Y. Yemez, Three-Dimensional Image Processing (3DIP) and Applications. Edited by Baskurt Atilla M. Proceedings of the SPIE, Volume 7526, pp. 75260M-75260M-12 (2010).
15. P. Scovanner, S. Ali, M. Shah, A 3-Dimensional SIFT Descriptor and its Application to Action Recognition, Poster, Proc. ACM Multimedia, (2007).
16. W. Cheung G. Hamarneh, n-SIFT: n-dimensional Scale Invariant Feature Transform for Matching Medical Images. Proc. IEEE ISBI, pp. 720-723, (2007).
17. V. Jain, H. Zhang. A Spectral Approach to Shape-Based Retrieval of Articulated 3D Models, Computer-Aided Design, Vol. 39, Issue 5, pp. 398-407, 2007
18. 3D-Voxelizer, by Carlos Martinez-Ortiz   http://www.mathworks.com/matlabcentral/fileexchange/21044-3d-voxel
19. P. Shilane, P. Min, M. Kazhdan, and T. Funkhouser. The Princeton Shape Benchmark. Shape Modeling International, June 2004.\
20. K. Siddiqi K., J. Zhang, D. Maxrini, A. Shokoufandeh, S. Bouix, D S. Dickinson: Retrieving articulated 3D models using medial surfaces. Machine Vision and Applications 19, 4 (2008), 261–274.
21. Z. Lian, A. Godil, T. Fabry, T. Furuya, J. Hermans, R. Ohbuchi, C. Shu, D. Smeets, P. Suetens, D. Vandermeulen, S. Wuhrer, 2010. SHREC'10 Track: Non-rigid 3D Shape Retrieval in Proceedings of the Eurographics/ACM SIGGRAPH Symposium on 3D Object Retrieval

22. Z. Lian, A. Godil, X. Sun. Visual Similarity Based 3D Shape Retrieval Using Bag-of-Features. In Proceedings of Shape Modeling International (2010). pp.25-36.
23. B. Bustos, A.D. Keim, D. Saupe, T. Schreck, V. D. Vranic. Feature-based similarity search in 3d object databases. ACM Computing Surveys 37(4), 345–387, 2005